\documentclass[
     sigconf = true,    
     screen = true,     
     review = false,     
     anonymous = false,  
]{acmart}

\usepackage{hyperref}
\usepackage{url}
\usepackage{booktabs,subcaption,amsfonts,dcolumn}
\usepackage{comment}
\AtBeginDocument{%
  \providecommand\BibTeX{{%
    \normalfont B\kern-0.5em{\scshape i\kern-0.25em b}\kern-0.8em\TeX}}}

\copyrightyear{2022}
\acmYear{2022}
\setcopyright{acmlicensed}
\acmConference[GECCO '22]{Genetic and
Evolutionary Computation Conference}{July 9--13, 2022}{Boston, MA, USA}
\acmBooktitle{Genetic and Evolutionary Computation Conference (GECCO '22),
July 9--13, 2022, Boston, MA, USA}
\acmPrice{15.00}
\acmDOI{10.1145/3512290.3528809}
\acmISBN{978-1-4503-9237-2/22/07}

\newcommand{\carola}[1]{\textbf{\textcolor{red}{Carola: #1}}}

\newcommand{\squeezeup}{\vspace{-1.5mm}}
\begin{document}

\title{SELECTOR: Selecting a Representative Benchmark Suite for Reproducible Statistical Comparison}


\author{Gjorgjina Cenikj}
\affiliation{
  \institution{Computer Systems Department\\ Jo\v{z}ef Stefan Institute}
  \country{Slovenia}
}
\email{gjorgjina.cenikj@ijs.si}

\author{Ryan Dieter Lang}
\affiliation{
  \institution{Computer Science Division \\ Stellenbosch University}
  \country{South Africa}
}
\email{langr@sun.ac.za}

\author{Andries Petrus Engelbrecht}
\affiliation{
  \institution{Department of Industrial Engineering, and Computer Science Division \\ Stellenbosch University}
  \country{South Africa}
}
\email{engel@sun.ac.za}

\author{Carola Doerr} 
\affiliation{%
  \institution{Sorbonne Universit\'e\\ CNRS, LIP6}
  \streetaddress{}
  \city{Paris} 
  \country{France}}
\email{carola.doerr@lip6.fr}

\author{Peter Koro\v{s}ec}
\affiliation{
  \institution{Computer Systems Department\\ Jo\v{z}ef Stefan Institute}
  \country{Slovenia}
}
\email{peter.korosec@ijs.si}

\author{Tome Eftimov}
\affiliation{
  \institution{Computer Systems Department\\ Jo\v{z}ef Stefan Institute}
  \country{Slovenia}
}
\email{tome.eftimov@ijs.si}

\begin{abstract}
Fair algorithm evaluation is conditioned on the existence of high-quality benchmark datasets that are non-redundant and are representative of typical optimization scenarios.
In this paper, we evaluate three heuristics for selecting diverse problem instances which should be involved in the comparison of optimization algorithms in order to ensure robust statistical algorithm performance analysis. The first approach employs clustering to identify similar groups of problem instances and subsequent sampling from each cluster to construct new benchmarks, while the other two approaches use graph algorithms for identifying dominating and maximal independent sets of nodes.
We demonstrate the applicability of the proposed heuristics by performing a statistical performance analysis of five portfolios consisting of three optimization algorithms on five of the most commonly used optimization benchmarks.

The results indicate that the statistical analyses of the algorithms' performance, conducted on each benchmark separately, produce conflicting outcomes, which can be used to give a false indication of the superiority of one algorithm over another. On the other hand, when the analysis is conducted on the problem instances selected with the proposed heuristics, which uniformly cover the problem landscape, the statistical outcomes are robust and consistent.



\end{abstract}



\maketitle

\sloppy 

\squeezeup
\section{Introduction}

Reliable and unbiased algorithm performance evaluation plays an indispensable role in the analysis of the strengths and weaknesses of existing algorithms, tracking improvements, and establishing future research efforts~\cite{overfitting_model_selection_bias}. 



The main task of benchmarking in evolutionary computation involves evaluation of the performance of an algorithm against other algorithms. To do this, three main questions should be taken with great care: i) which problem instances should be selected for the comparison study, ii) how to design the experiments that lead to reproducible and replicable results, and iii) which performance measures should be analyzed and with which statistical approach. 

Even though great progress has been achieved by proposing more robust statistical methodologies for benchmarking, those approaches focused only on the third question of the benchmarking theory, with the assumption that the first and the second one are already done with great care. However, the selection of the problem instances (i.e., the benchmark suite) that will be included in the analysis can have a huge impact on the experimental design and statistical analysis performed using the performance data. 
It can happen that the same algorithm portfolio (i.e., set of algorithm instances) evaluated using some statistical methodology on different sets of problem instances results in different winning algorithms. This means that the selection of the problem instances can lead to biased performance analysis (i.e., selecting problem instances in favor of the winning algorithm). 
This allows researchers to present results that make their algorithm look superior to the others. 

In the field of single-objective optimization, algorithm performance is commonly assessed experimentally, for example on the benchmarks provided by the IEEE Congress on Evolutionary Computation (CEC) Special Sessions and Competitions on Real-Parameter Single-Objective Optimization ~\cite{cec2013, cec2014, cec2015, cec2017} and the Genetic and Evolutionary Computation Conference (GECCO) Black-box Optimization Benchmarking (BBOB) ~\cite{bbob,hansen2020coco} workshops. 
However, recent studies suggest that the CEC and the BBOB benchmark suites may provide poor coverage of the space of numerical optimization problems~\cite{lacroix2019, urban_complementary_analysis, munoz_generating_new_instances}.
Even more, it has also been shown that the benchmark problem classes are highly correlated from the perspective of the performance space~\cite{zhang2019similarity,christie2018investigating}.

In this paper, we address the selection of problem instances which should be involved in the comparison of optimization algorithms in order to ensure robust statistical algorithm performance analysis. 

One approach to select the problem instances is the Instance Space Analysis (ISA) methodology~\cite{smith2014towards}, which uses visualization to assess the effect of instance characteristics on algorithm performance, by finding areas in the problem landscape space where some algorithms perform better than the others. The idea is to select instances that maximize the performance difference between algorithms to highlight their strengths and weaknesses~\cite{isa_optimization,isa_optimization_moo}. 
Racing has also been explored for the selection of problems that highlight performance differences between algorithms~\cite{racing_problem_selection}.
An alternative approach was proposed in~\cite{Weise20ASOCwmodel} where instances are selected based on clustering applied in \emph{performance space,} i.e., they were selected to maximize the diversity of the comparisons that one could perform between the algorithms. While our approach has a stronger motivation in selecting instances for robust comparisons, the approach in~\cite{Weise20ASOCwmodel} has a stronger motivation in benchmarking for algorithm analysis.

Other studies explore the diversity of single-objective optimization problems~\cite{vskvorc2020understanding,lang2021exploratory,meunier2021black}, without focusing on selection heuristics that can produce more comprehensive benchmark datasets.

The main difference of our approach with the aforementioned approaches is that the selection is done based on landscape characteristics of the problem instances, without investigating the relations with the performance space. Consequently, our approach is not affected by the selection of the algorithm portfolio which will be included in the analysis and it enables the selection of diverse instances based on the landscape characteristics, regardless of which optimization algorithms will be further run and compared.

\textbf{Our contribution:} 
We evaluate three approaches for the selection of problem instances, which should be involved in the comparison of optimization algorithms in order to ensure robust statistical algorithm performance analysis, all of which first involve generating a numerical representation of the problem instances using exploratory landscape analysis (ELA).
The first approach then employs clustering to identify similar groups of problem instances and subsequent uniform sampling from each cluster to construct new benchmarks. The other two approaches are based on graph theory algorithms for finding a dominating set (DS) and a maximal independent set (MIS) of nodes in a graph.

We show that the results of the statistical algorithm performance analysis executed on a single benchmark (e.g., BBOB, CEC2013, CEC2014, CEC2015, and CEC2017) are not consistent when a different benchmark is being used. Such results allow researchers to select the benchmark where their algorithm is superior, which are bias to the researchers' preference. On the other hand, the subsets of problem instances produced by the proposed approaches provide a robust, consistent, and reproducible statistical analysis.

\textbf{Outline:} 
The reminder of the paper is organized as follows: Section~\ref{sec:background} provides the background, Section~\ref{sec:methodolgy} presents three different sampling heuristics to select problem instances that guarantee reproducible statistical outcomes. Our key results are presented in Section~\ref{sec:results}. We conclude the paper in Section~\ref{sec:conclusion}.

\textbf{Reproducibility:} The performance data as well as the landscape data used in this study are available online at \url{https://zenodo.org/record/5940558}.
The code is available at \url{https://anonymous.4open.science/r/SELECTOR-80F1}.

\section{Background}
\label{sec:background}
In this section, we present two core components of our analysis: exploratory landscape analysis (Section~\ref{ssec:ela}, used to characterize the problem instances via numerical representations) and statistical performance assessment (Section~\ref{sec:performance}, used to compare the algorithms).

\subsection{Exploratory Landscape Analysis}
\label{ssec:ela}

Exploratory landscape analysis (ELA)~\cite{mersmann2011exploratory} is an approach to characterize black-box optimization problem instances via numerical measures (\emph{features}) that each describe a different aspect of the problem instances.
A convenient way to compute ELA features is provided by the \textit{flacco} R package~\cite{KerschkeT2019flacco}.
This packages offers 343 different feature values split into 17 features groups, including dispersion, information content, meta model, nearest better clustering, principal component analysis.

\subsection{Performance Assessment} 
\label{sec:performance}
Once the benchmark problems are selected, the performance assessment can be done in a fixed-budget or fixed-target scenario.
The fixed-budget scenario measures the quality of the solution achieved with a given computational budget, while the fixed-target scenario returns the budget required to achieve a certain target. 

The quality of a solution in single-objective optimization can be expressed in terms of absolute fitness value or in terms of \emph{target precision}, i.e., the difference of this absolute fitness to that of an optimal solution. 
After selecting the performance measure, statistical analyses play an important role in the interpretation of the obtained results. 
A common practice is use a non-parametric test on the mean results obtained from multiple runs of each algorithm on each problem instance~\cite{garcia2009study}. Recent advances involve the use of Deep Statistical Comparisons (DSC)~\cite{eftimov2017novel}. In comparison to using individual descriptive statistics, such as the mean or median, from multiple independent runs on a benchmark problem, DSC is based on the whole distribution of the independent runs, with the goal to ensure a more robust statistical analysis, as the mean can be affected by outliers (i.e., poor runs) and the medians can be in some small $\epsilon$-neighbourhood of the performance space, which is practically insignificant. 

\section{Methodology}
\label{sec:methodolgy}

\begin{figure}
    \centering
    \includegraphics[width=\linewidth]{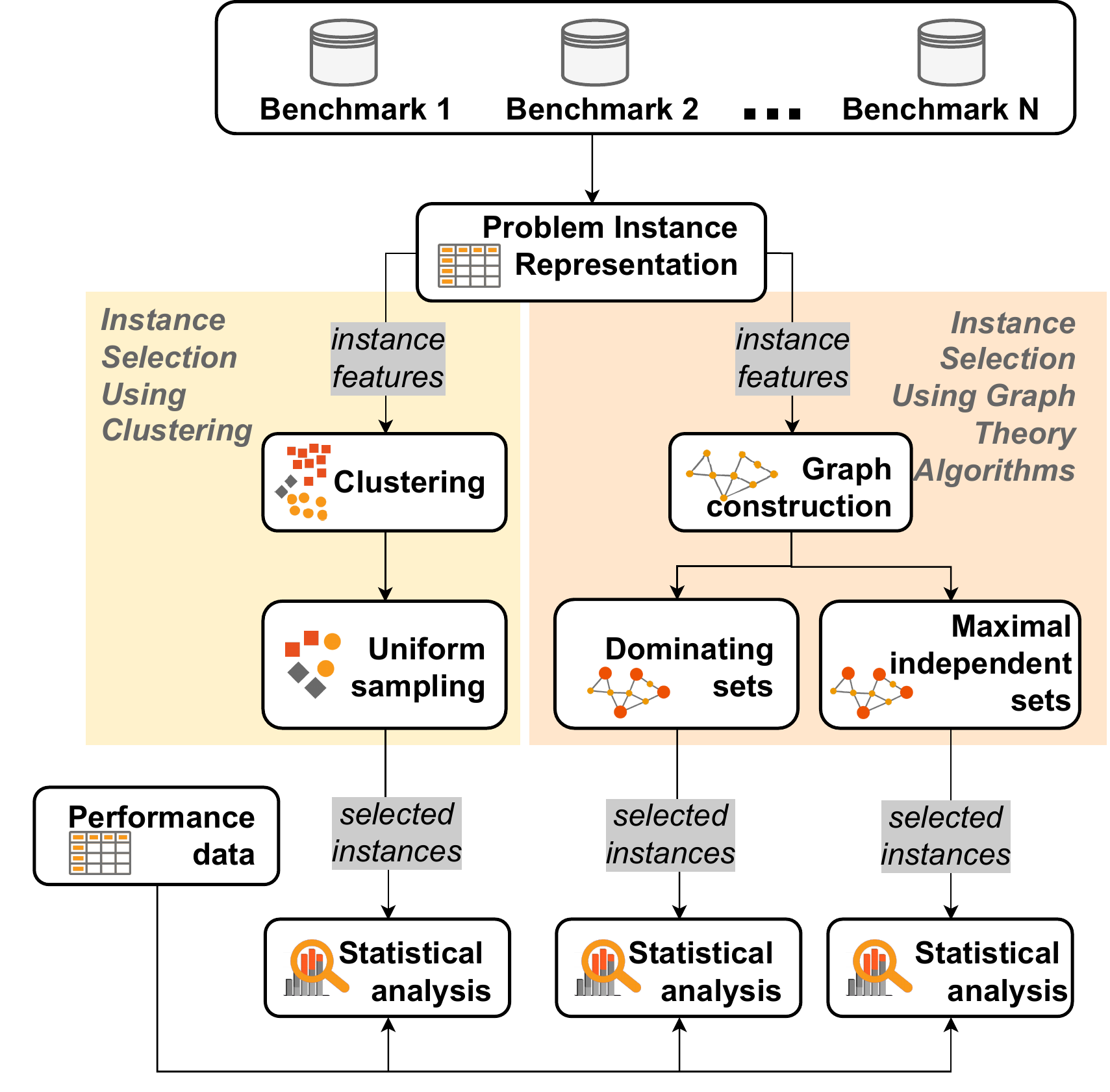}
    \caption{Overview of the proposed methodology.}
    \label{fig:methodology_overview}
\end{figure}

Figure~\ref{fig:methodology_overview} gives an overview of the proposed methodology. Given a set of benchmarks, the first step involves generating a vectorized representation for each problem instance using ELA features. 

The heuristic which performs instance selection using clustering then proceeds to cluster the representations of all problem instances and perform sampling to identify a set of instances which will uniformly cover the problem landscape.
On the other hand, the heuristics which perform instance selection using graph theory algorithms construct a graph based on the similarity of the problem instances, and identify a subset of instances on which the algorithms should be compared, by applying the algorithms for finding DS and MIS in the graph.
Once the instances are selected by each heuristic, a statistical algorithm performance analysis is conducted by comparing the algorithms' performance on the instances selected by each heuristic independently. We further show that the selected instances from each heuristic enable a consistent and robust statistical analysis, as opposed to using a single benchmark.

\subsection{Problem Instance Representation}

We extract the ELA features using flacco's implementation of the improved Latin Hypercube sampling technique~\cite{iLHS}. We use a sample size of $N = 800 \times D = 8000$, because it was shown in \cite{lang2021exploratory} that this sample size enables robust generation/calculation of ELA features. 

As mentioned in Section~\ref{ssec:ela}, \textit{flacco} provides a number of ELA feature sets.
In this study, a total of \textbf{64 features} are included: 
16 dispersion measures~\cite{lunacek_dispersion_2006}, 
three y-distribution measures~\cite{mersmann2011exploratory}, 
18 level-set measures~\cite{mersmann2011exploratory}, 
nine meta-model measures~\cite{mersmann2011exploratory}, 
five information content measures~\cite{munoz_exploratory_2015}, 
five nearest better clustering measures~\cite{kerschke_detecting_2015}, 
and eight principal component analysis measures~\cite{KerschkeT2019flacco}. 
These ELA features have been selected, as they do not require additional sampling beyond the initial design. 
Out of these features, the \textit{ic.eps.s} feature is removed, as it contains null values for some of the instances.
The features are calculated 30 times, and each problem instance is represented by a 63-dimensional numerical vector, obtained as the median of the features obtained in the 30 runs of the feature calculation.


\subsection{Instance Selection Using Clustering}

Hierarchical agglomerative clustering is used to partition the entire set of problem instances. 
The problem instances are represented using the previously described ELA features, and cosine similarity of the ELA features is used to calculate the similarity of two problem instances. 
Cosine similarity is used as it is dependent only on the direction of vectors and not their length, rendering them less sensitive to outliers compared to other similarity measures~\citep{shirkhorshidi2015comparison}. 

We selected the hierarchical clustering, since in k-means clustering one starts with a random choice of clusters, and the results produced by running the algorithm many times may differ. Hierarchical clustering thus ensures better
reproducibility of our results. 

To estimate the number of clusters we used the silhouette score. In addition, when determining the number of clusters and the number of instances that will be sampled from each cluster, one has to make sure that the total number of sampled instances will be enough to perform the statistical analysis.

To construct an unbiased benchmark set of problem instances that will provide a more uniform coverage of the feature space, we sample an equal number of problem instances from each cluster to make each of the landscape they cover represented with the same number of instances. To see if the sampling ensures reproducible statistical outcome, we should repeat the sampling several times and perform the statistical analysis.





\subsection{Instance Selection Using Graph Theory Algorithms}

In order to select diverse instances from all of the instances in the benchmarks, we explore two graph theory algorithms, which involve finding the dominating and independent sets of nodes in a graph, respectively.

\subsubsection{Definition of Dominating and Maximal Independent Sets}

Given an undirected, acyclic, finite graph $G$ with no multiple edges, represented with a set of vertices $V$ and a set of edges $E$, one can define a dominating set (DS), $D$, as a subset of the nodes in $G$, such that every node not in $D$ is adjacent to at least one member of $D$~\cite{definitions_dom}. 
Formally, a dominating set $D$ is a subset of $V$ such that for all 
$v \in V - D$ 
its neighborhood set $N(v)$ has non-empty intersection with $D$, i.e., $N(v) \cap D \neq \emptyset$.

On the other hand, an independent set $I$ is a set of nodes such that the subgraph of $G$ induced by these nodes contains no edges, i.e., none of the nodes in $I$ are adjacent to each other in $G$, or formally, $\forall u, v \in I, N(v) \cap \{u\} = \emptyset $. A maximal independent set (MIS) is then defined as an independent set such that no node can be added to it without
violating independence~\cite{definitions_mis}.


\subsubsection{Application of Dominating and Maximal Independent Sets Algorithms for Benchmark Sampling}

The first step in our proposed methodology based on graph theory involves the construction of a homogeneous graph where the nodes represent problem instances, and an edge between two nodes indicates that they are similar according to some similarity measure.

We define a pair of problem instances to be similar if the cosine similarity of the ELA features used to represent the two instances exceeds a certain threshold. For this purpose, we evaluate several values for the similarity threshold. 

When the graph is constructed in such a manner,
the application of the DS and MIS algorithms produces a subset of diverse problem instances from the considered benchmarks. Both algorithms ensure that instances which are not similar to any of the other instances (nodes with no neighbours in the graph) will be part of the extracted set. The DS algorithm selects instances so that all the instances which are not selected, are similar to at least one of the selected instances, while the MIS algorithm selects instances which are not similar to each other.

For our experiments, we use the implementation of the DS and MIS algorithms provided by the \textit{networkx} python library~\cite{networkx}. Since these algorithms are stochastic in nature (in particular due to the random selection of the initial node set), we repeat the experiment 30 times with different random seeds, and each algorithm produces 30 different sets of problem instances.

\subsection{Statistical Analysis} \label{statistical analysis}
The sets of problem instances produced with both the clustering and the graph theory approaches are evaluated in terms of the robustness and consistency of the statistical algorithm performance analysis which can be executed when the algorithms are compared on the performance they achieve on these instances.

The Deep Statistical Comparison (DSC) raking scheme~\cite{eftimov2017novel} is used to rank the algorithms on each problem instance selected in the benchmark suite. 
The rankings determined by DSC are analyzed with the Friedman non-parametric statistical test.
The null hypothesis of the Friedman test states that all of the investigated algorithms achieve equivalent performance.

If the null hypothesis of the Friedman test is rejected, i.e. the algorithm performances are not equivalent, the Nemenyi post-hoc test~\cite{nemenyi} is used to identify the algorithm pairs which provide statistically different performance (i.e., multi-hypothesis correction scenario performed for each selected benchmark suite)~\cite{demsar_statistics}. 
To determine the robustness of the evaluation done on the benchmarks produced in each of the 30 different executions of the sampling process in the clustering approach and the 30 different executions of the DS and MIS algorithms, we keep track of the number of times (out of 30) that the Nemenyi test indicates that each algorithm pair produces an equivalent performance. The counting approach is only an indicator if the statistical results are robust if we repeat it using different problem instances selected by the proposed heuristics. 

The Nemenyi test provides a $p$-value for each pair of algorithms. 
If the $p$-value is greater or equal than the significance level, the null hypothesis is not rejected, and  (i.e., we translate this to 1, no statistical significance between the performance of the algorithms), otherwise there is a statistical significance between the performance of the compared pair of algorithms (i.e., we translate this to 0, a statistical significance is detected).
A significance level of 0.05 is used in the experimental procedure.

We use the implementation of the Friedman test provided by the \textit{scmamp} R package ~\cite{scmamp} (version 0.2.55) and the Nemenyi post-hoc test from the \textit{PMCMR} R package ~\cite{pmcmr} (version 4.3).

\section{Results and Discussion}
\label{sec:results}

This section presents the data used to evaluate the proposed approach, followed by the statistical results obtained when selecting a representative benchmark suite.

\subsection{Data}

The benchmark suites used in this study are a subset of those utilized in~\cite{lang2021exploratory}.
Specifically, the BBOB~\cite{bbob,hansen2020coco} and CEC~\cite{cec2013, cec2014, cec2015, cec2017} benchmark suites from the 2013, 2014, 2015, and 2017 single-objective competitions are used.
The BBOB benchmark suite defines 24 base problems, from which different instances can be generated by applying transformations, such as rotation and scaling.
The first five instances of the BBOB functions are used, resulting in 120 benchmark problems.
The CEC2013 benchmark suite contains 28 problems, CEC2014 contains 30 problems, CEC2015 contains 15 problems, and CEC2017 contains 29 problems. All CEC problems are included only with a single instance. 
Therefore, a total of 222 benchmark problems are investigated.
For the experimental procedure, the dimensionality of the decision variable space is set to 10.

To illustrate the rankings of optimization algorithms on the benchmark problems, an algorithm portfolio of three single-objective optimization algorithms is selected: 
the Covariance Matrix Adaption Evolutionary Strategy (CMA-ES)~\cite{hansen2001self_adaptation_es}, 
RealSpace Particle Swarm Optimization (RSPSO)~\cite{kennedy1995pso},
and Differential Evolution (DE)~\cite{storn1997differential}.
This algorithm portfolio was chosen somewhat arbitrarily, as the key interest in this study is in presenting the general landscape-aware instance selection pipeline itself, and in analyzing whether it can provide reproducible statistical outcomes rather than an in-depth study of any particular algorithm portfolio. For validation of the results, we have repeated the experiments for four additional algorithm portfolios of three randomly selected algorithms (results/tables are available in our GitHub repository). 

The \textit{Nevergrad}~\cite{nevergrad_github} library is used for its implementations of the optimization algorithms with the default hyperparameter configurations provided in the library.
All algorithms are run for a fixed budget of 100,000 function evaluations.
An algorithm run is terminated either if the function evaluation budget is exhausted, or if the algorithm reaches within $\epsilon = 10^{-8}$ of the function's known global minimum.
Due to the stochastic nature of the investigated algorithms, each algorithm is run for 30 independent runs on each problem instance.

\subsection{Benchmarking using already established benchmark suites}
The best practices of comparing the performance of a newly developed algorithm involve using an already established benchmark suite. However, one of the issues is which benchmark suite to be involved, since some of them can be in favour of the newly developed algorithm. To show the difference between different benchmark suites, we compare the three algorithms separately on each benchmark suite. The results from the comparison, after the Nemenyi post-hoc test, are presented in Table~\ref{tab:benchmarks_merged}.
We report the raw p-values, along with a binary indicator of statistical significance, which has a value of 1 when no statistical significance between the difference in performance of the algorithms has been found, and a value of 0, when a statistical significance is detected.

%

\begin{table*}[th]
\caption{Statistical comparison of the three algorithms on already established benchmark suites. }
\label{tab:benchmarks_merged}
\begin{tabular}{|r|r|r|r|r|r|r|r|r|r|r|r|r|r|r|}
  \hline
  & \multicolumn{2}{c|}{BBOB} 
  & \multicolumn{2}{c|}{CEC 2013} 
  & \multicolumn{2}{c|}{CEC 2014} 
  & \multicolumn{2}{c|}{CEC 2015} 
  & \multicolumn{2}{c|}{CEC 2017}
  & \multicolumn{2}{c|}{All}\\
\hline
 & DE & RSPSO
 & DE & RSPSO
 & DE & RSPSO
 & DE & RSPSO
 & DE & RSPSO
 & DE & RSPSO\\ 
  \hline
  RSPSO & 0.00/0 & &  0.95/1 &  & 0.29/1 &  &  0.02/0 & &  0.00/0 &  & 0.00/0 & \\ 
  CMA & 0.33/1 & 0.00/0 & 0.47/1 & 0.65/1 & 0.75/1 & 0.07/1 & 0.04/0 & 0.98/1 & 0.97/1 & 0.00/0 & 0.06/1 & 0.00/0\\ 
   \hline
\end{tabular}
\end{table*}

Using the obtained statistical outcomes across the benchmark suites, it is obvious that different statistical outcomes are produced. 

On the BBOB benchmark suite, there is no statistical significance between the performance of the CMA and DE, however their performances differ statistically significantly from the performance of RSPSO. 
When applying the mean DSC ranking on the BBOB and CEC2017 benchmarks, the results suggest that RSPSO is better than CMA and DE, while no statistical significance is found between the performance of CMA and DE.

On the CEC2013 and CEC2014 benchmark suites, there is no statistical significance between the performance of any pair of algorithms. 
The results obtained on CEC2015 benchmark suite indicate that there is a statistically significant difference between the performance of the pairs of algorithms (DE, RSPSO) and (DE, CMA), while there is no statistical significance between CMA and RSPSO. 

Using all problem instances from all benchmark suites, the RSPSO is the superior algorithm, and there is no statistical difference between the performances of CMA and DE. 

Looking at the results, the crucial question is what one wants to show as the outcome of the statistical comparison. One option is to show that DE is superior and for this purpose we could select CEC2015 benchmark suite. Another option is to show that all algorithms have 
hardly distinguishable performance, by selecting the CEC2013 or the CEC2014 suite. A third option is to select BBOB or CEC2017 and show that RSPSO is the superior one. 
This kind of manipulation with the results allows publishing biased statistical results or presenting results based on the researchers' preference.

To overcome this issue, we propose three heuristics for selecting problem instances that provide reproducible statistical outcomes.

\subsection{Instance Selection Using Clustering}
To cluster the problem instances, agglomerate hierarchical clustering was applied using the \textit{scikit-learn} python library~\cite{scikit-learn}.
We estimate the number of clusters to be 12, based on the silhouette score and the fact that a minimum of 10 samples are needed to execute the statistical analysis.

\begin{figure}
    \centering
    \begin{subfigure}[b]{\linewidth}
    \includegraphics[width=\linewidth]{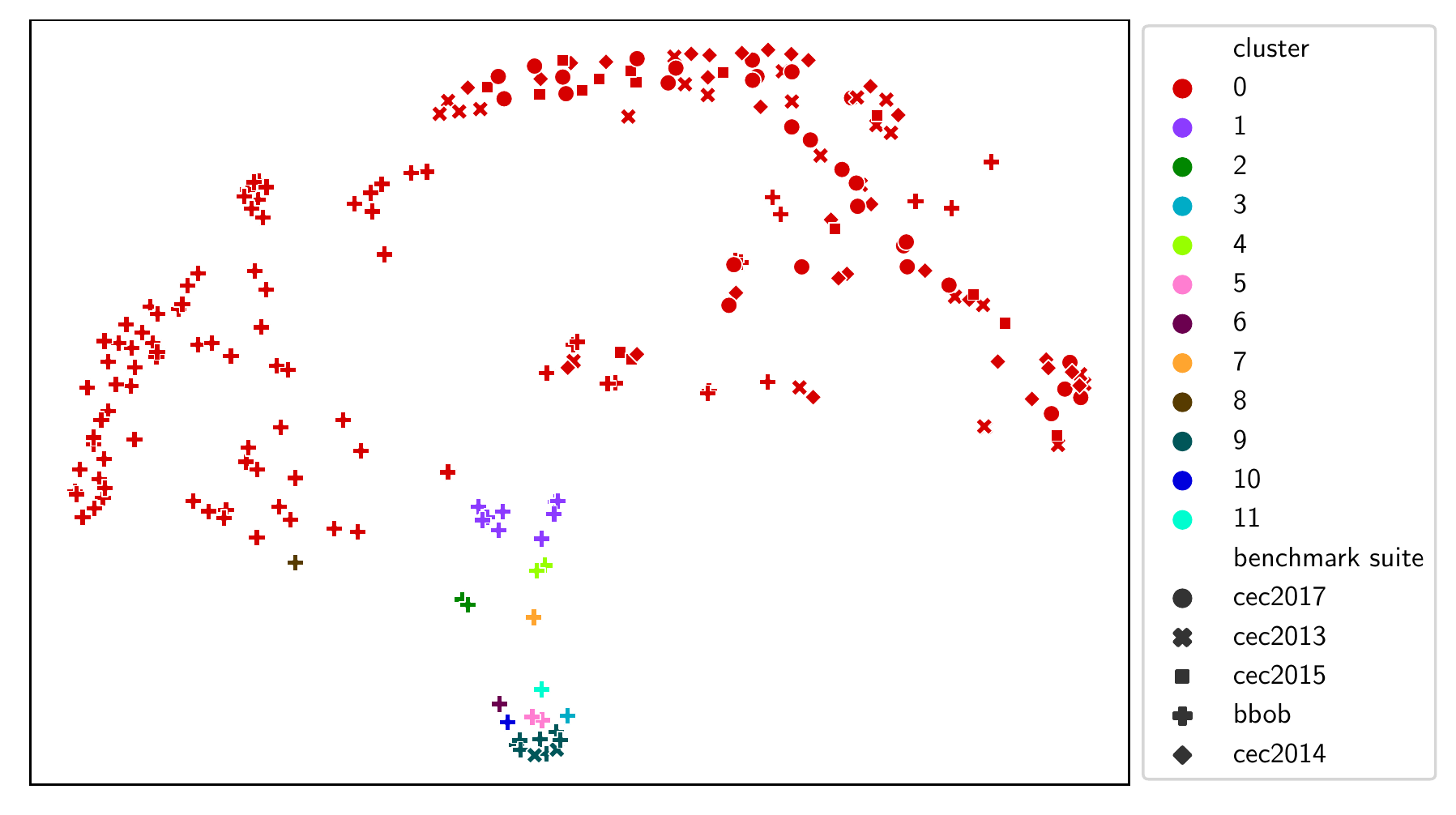}
    \caption{Visualization of the initial 12 clusters}
    \label{fig:first_part_clustering}
    \end{subfigure}
    \hfill
    
\begin{subfigure}[b]{\linewidth}
    \centering
    \includegraphics[width=\linewidth]{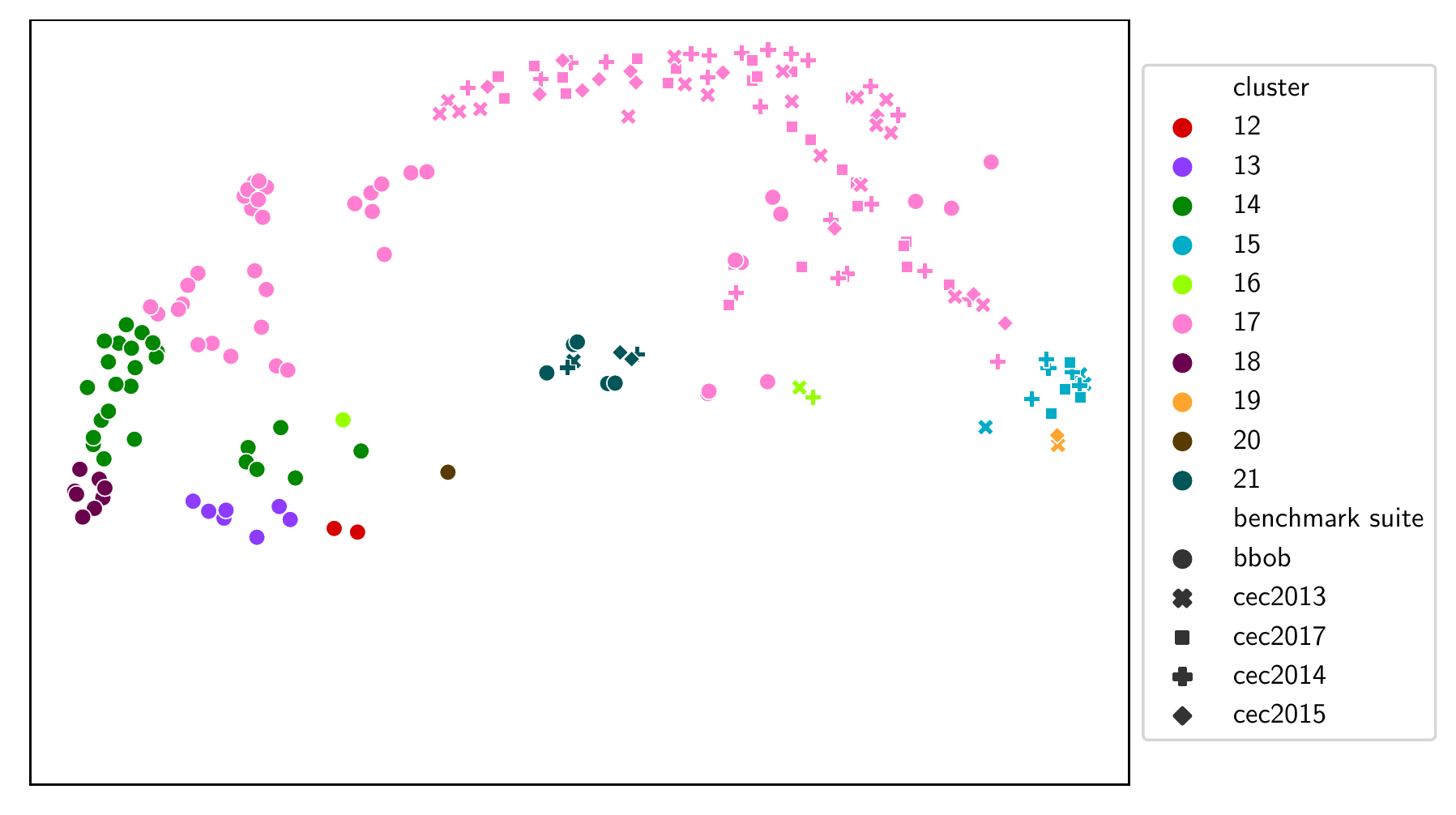}
    \caption{Visualization of the subclusters obtained by re-clustering the largest of the 12 clusters}
    \label{fig:second_part_clustering}
\end{subfigure}
    
    \caption{Visualization of the clustering, obtained by projecting the ELA instance features into 2 dimensions using t-distributed stochastic neighbor embedding with cosine similarity.  Different colors indicate different clusters, while different shapes indicate different benchmarks.}
\end{figure}

\begin{figure}
    \centering
    \includegraphics[width=0.8\linewidth]{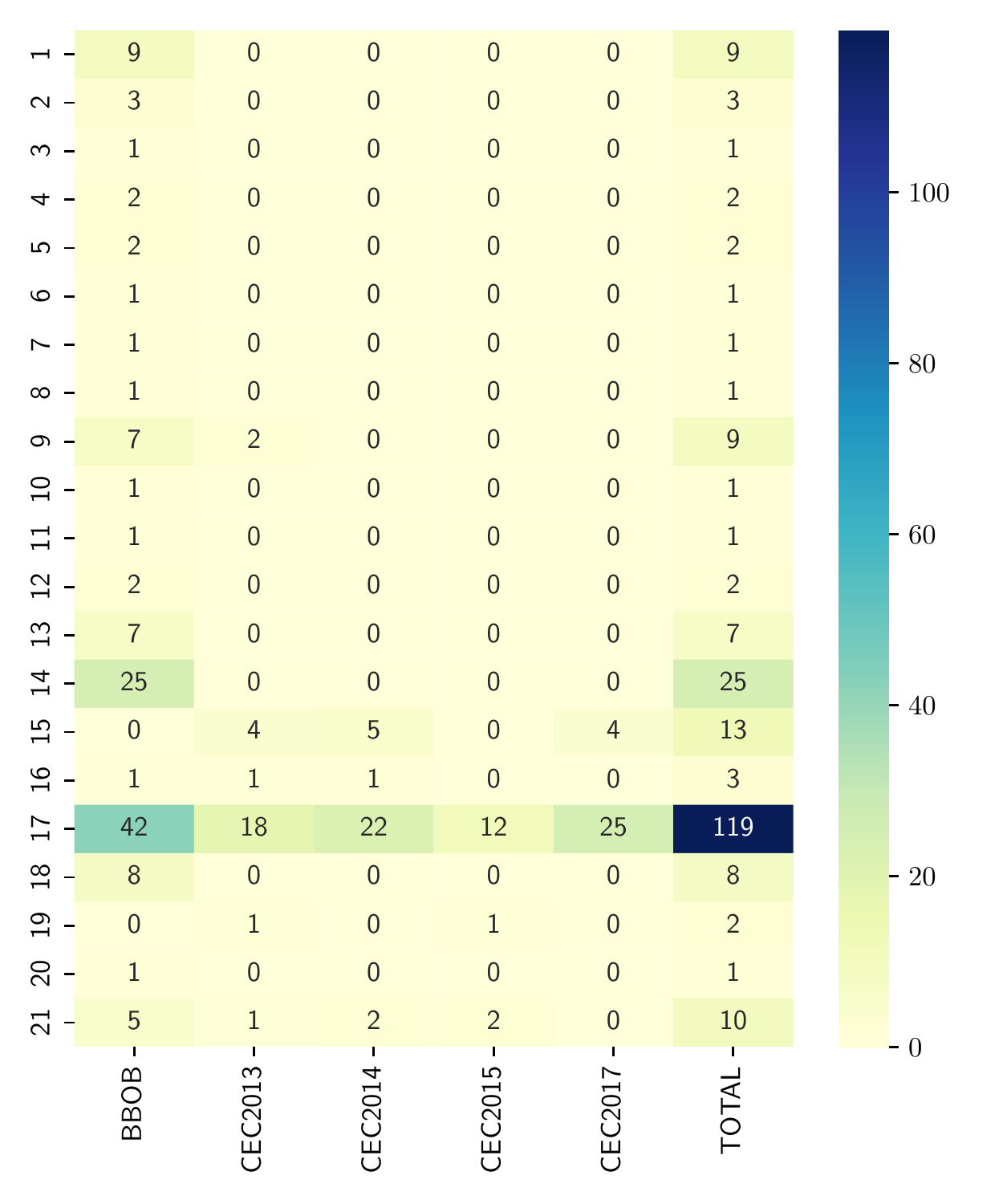}
    \caption{Number of instances from each benchmark suite, and total number of instances in each cluster. }
    \label{fig:benchmarks_per_cluster_heatmap}
\end{figure}

Figure~\ref{fig:first_part_clustering} visualizes these 12 clusters, as computed by the agglomerate hierarchical clustering approach.
We easily spot that the number of problem instances per cluster is not balanced: There are six clusters that consist of a single instance, while the largest cluster contains 190 instances. 
We therefore took the instances that belong to the largest cluster and we further clustered them to find subclusters. Looking at the silhouette scores obtained when the large cluster is broken up, we ended up with 10 subclusters, leaving us with \textbf{a total number of 21 clusters.} These clusters are visualized in Figure~\ref{fig:second_part_clustering} and Figure~\ref{fig:benchmarks_per_cluster_heatmap} features the number of instances from each benchmark suite in each cluster. 

As can be seen from Figure~\ref{fig:benchmarks_per_cluster_heatmap},
the BBOB benchmark covers 19 out of the 21 clusters with at least one instance. On the other hand, the CEC2013, CEC2014, CEC2015 and CEC2017 benchmarks cover 6, 4, 3, and 2 clusters, respectively, and are mostly distributed in the subclusters obtained with the second clustering. This suggests that the BBOB benchmark covers more of the landscape, however it should be combined with some of the CEC benchmark suites to cover a larger portion of the problem landscape.

From Figure~\ref{fig:benchmarks_per_cluster_heatmap}, we can see that there are six clusters with a single problem instance, indicating that we need to sample one instance from each cluster. The question that arises here is which instance should be selected from the larger clusters. If we select the instance randomly, it can be closer to some of the other clusters and will not represent the cluster structure. 
For this purpose, we need to find the problem instance that is closest to the centroid of each cluster. To calculate the centroids, we use all instances that belong to each cluster and calculate the mean value of each ELA feature. Next, using cosine similarity we find the closest problem instance to the centroid of each cluster. These problem instances are selected as representatives and further involved in the benchmark suite.

The results of comparing the algorithms using the Friedman and Nemenyi post-hoc tests on the selected problem instances are presented on the left of Table~\ref{tab:21}, where we can see that only RSPSO and CMA have a statistically significant performance difference.

\begin{table}[ht]
\centering
\caption{Results of the Friedman test and the Nemenyi post-hoc test for the statistical comparison of the three algorithms using the benchmark suites selected from the 21 and the 26 clusters, respectively.}
\label{tab:21}
\begin{tabular}{|r|r|r|r|r|}
  \hline
  & \multicolumn{2}{c|}{21 clusters} & \multicolumn{2}{c|}{26 clusters}\\
  \hline
 & DE & RSPSO  & DE & RSPSO \\ 
  \hline
  RSPSO & 0.24/1 & & 0.28/1 &   \\ 
  CMA & 0.48/1 & 0.02/0  & 0.51/1 & 0.02/0\\ 
   \hline
\end{tabular}
\end{table}

We repeated the experiment when the largest cluster is split into 15 clusters instead of 10, since the silhouette score is similar. In this case, we ended up with 26 clusters. The results of this experiment are reported on the right part of Table~\ref{tab:21}. They are consistent with those of the 21 clusters,   
i.e., statistical significance is reported only between the algorithms RSPSO and CMA.

Furthermore, we have also tested the statistical outcome when median values instead of mean values of the ELA features are used to select the centroids, but the same statistical outcome has been achieved as in the case with the mean values. 


To explore the sensitivity and flexibility of selecting the representatives for each cluster, instead of selecting the instance that is the closest to the cluster centroid, we can set a percentage of the instances closest to the cluster centroid that can be used as a representatives for the larger clusters. Furthermore, to create a benchmark suite, we should uniformly select one instance form the representatives per each cluster. To show if this leads to reproducible results, we repeat the selection several times, 15 (not 30 since we can not produce 30 different benchmark suites) and see if involving different representatives for the same cluster changes the statistical outcome. For this purpose, we have tested the following percentages: 12.5\%, and 25\%.
The statistical outcomes are presented in Table~\ref{tab:representative_ccluster}, where the results show that reproducible statistical outcomes are obtained no matter which percentage is selected to define the representative instances for the larger clusters. Even more, the results are statistically the same as the results obtained when the closest instance to the centroid from each cluster is selected and involved in the benchmark suite.
\begin{table}[ht]
\centering
\caption{Results of the Friedman test and the Nemenyi post-hoc test for the statistical comparison of the three algorithms using the benchmark suites selected by using different percentage of representatives for the larger clusters.}
\label{tab:representative_ccluster}
\begin{tabular}{|r|r|r|r|r|}
  \hline
  & \multicolumn{2}{c|}{12.5\% repres.} & \multicolumn{2}{c|}{25\% repres.}\\
  \hline
 & DE & RSPSO  & DE & RSPSO \\ 
  \hline
  RSPSO & 15.00 & & 15.00 &   \\ 
  CMA & 14.00 & 0.00  & 14.00 & 0.00\\ 
   \hline
\end{tabular}
\end{table}



\subsection{Instance Selection Using Graph Theory Algorithms}
Table~\ref{tab:graph_stats} displays the number of edges in the graphs produced with each similarity threshold, as well as the minimum, maximum and median of the number of problem instances produced by the DS and MIS algorithms in the 30 independent runs, respectively. As evident from the table, increasing the threshold results in a reduction of the number of edges in the graph, and an increase in the number of instances produced by the DS and MIS algorithms. 
Since the algorithms produce a maximum of 4 and 8 instances when run on the graphs constructed with a similarity threshold of 0.50 and 0.70, respectively, we do not consider these results in the statistical analysis, since they do not provide enough samples for the safe execution of the Friedman test.

\begin{table}[]
    \centering
    \caption{Descriptive statistics of the benchmark suites selected using the DS and MIS algorithms.}
    \label{tab:graph_stats}
   \begin{tabular}{|p{35px}|p{35px}|p{35px}|p{25px}|p{25px}|p{25px}|}
\hline
similarity threshold &  edge count & algorithm &  min &  max &   mean \\ \hline

0.50 & 20821 & DS &    2 &    4 &   3.30 \\
     &       & MIS &    3 &    4 &   3.37 \\ \hline
0.70 & 19940 & DS &    5 &    8 &   7.07 \\
     &       & MIS &    5 &    8 &   6.90 \\ \hline
0.90 & 19119 & DS &   11 &   12 &  11.43 \\
     &       & MIS &   11 &   13 &  11.47 \\ \hline
0.95 & 17460 & DS &   16 &   18 &  17.37 \\
     &       & MIS &   16 &   19 &  17.27 \\ \hline
0.97 & 15116 & DS &   20 &   24 &  22.13 \\
     &       & MIS &   21 &   24 &  22.63 \\ \hline

\end{tabular}
    
\end{table}

\begin{figure}
 \centering
    \includegraphics[width=\linewidth]{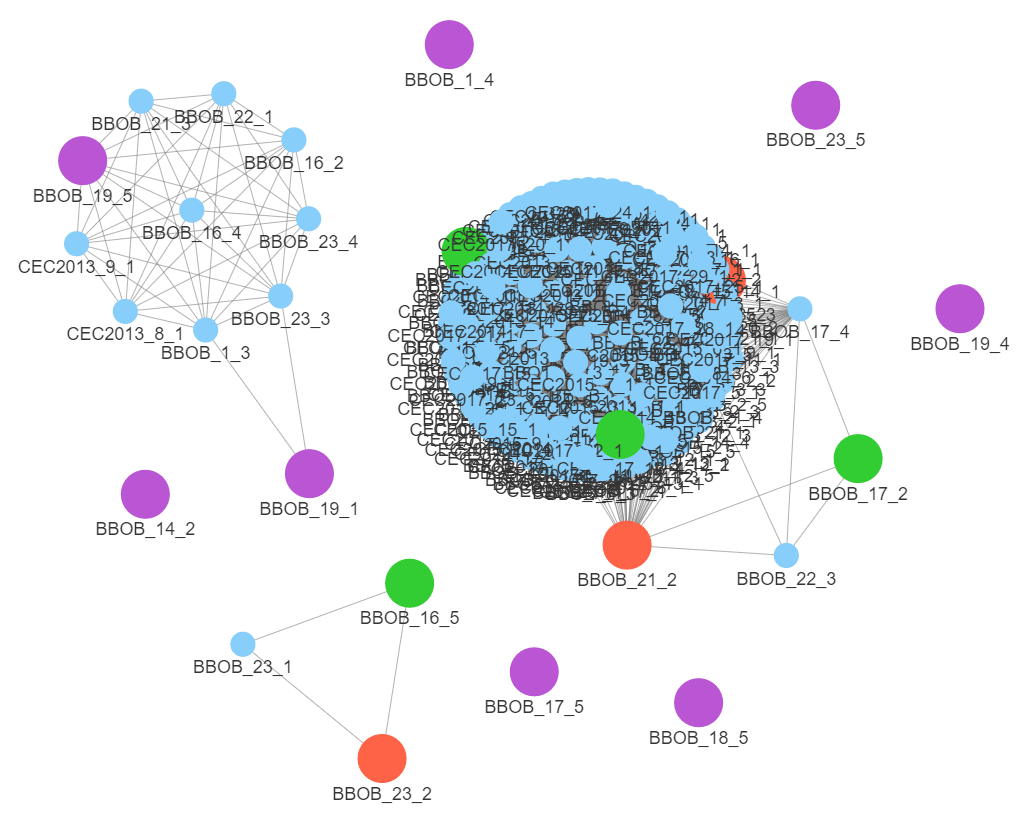}
    \caption{Visualization of the instances selected by the DS and MIS algorithms on the graph constructed with a minimum similarity threshold of 0.9.}
    \label{fig:graph_visualization}
\end{figure}

Figure~\ref{fig:graph_visualization} depicts the graph of instances constructed with a minimum similarity threshold of 0.9, meaning that each of the connected nodes have ELA features with a cosine similarity of at least 0.9. The orange nodes are the instances selected only by the DS algorithm, the green nodes are the instances selected by the MIS algorithm, while the purple nodes are the instances that were selected by both the MIS and the DS algorithm. 
The graph contains 9 connected components, with the 3 largest components containing 201, 11 and 3 nodes, respectively. There are also 6 zero-degree nodes (nodes without have any edges) that form an individual connected component. As can be seen from Figure~\ref{fig:graph_visualization}, all of these nodes represent instances from the BBOB benchmark. We can note that all of the zero-degree nodes are selected by both algorithms.

These solutions are based on only one run of the algorithm. The algorithms can produce different results depending on the initially selected nodes. For instance, if we take a look at the connected component of size 3, depicted in the lower left part of Figure~\ref{fig:graph_visualization}, in this run of the algorithms, the MIS algorithm selected the instance BBOB\_16\_5. In another run, the algorithm could select either one of the other two instances in the component, BBOB\_23\_1 or BBOB\_23\_2, however, it will ensure that two of these instances will never be selected together, which will prevent redundancy and encourage diversity in the benchmark.

\begin{figure}
    \centering
    \includegraphics[width=\linewidth]{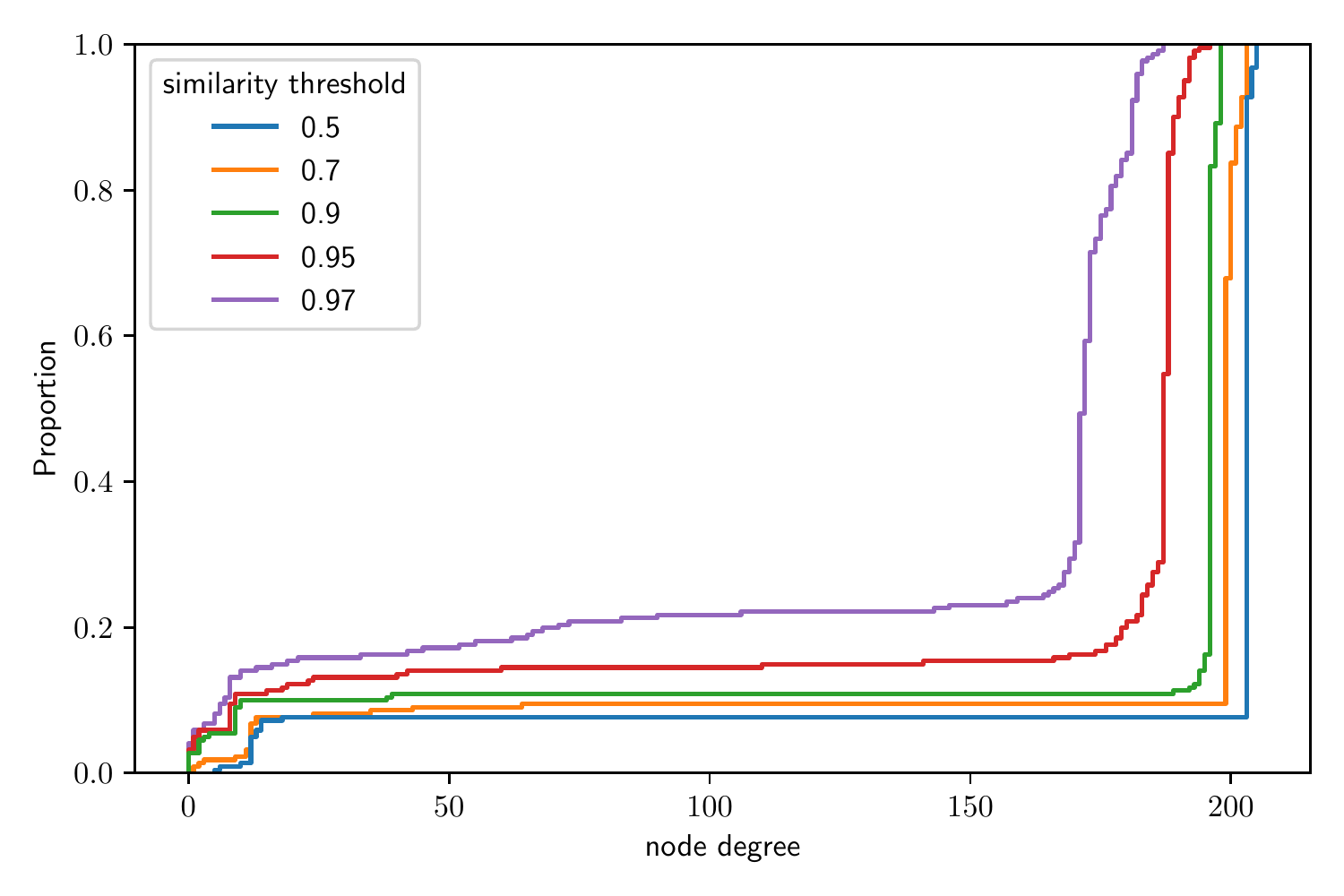}
    \caption{Empirical cumulative distribution plot of the node degrees for the graphs constructed with each similarity threshold, indicating the proportion of nodes having a node degree falling below each value on the x-axis.}
    \label{fig:node_degree_distribution}
\end{figure}

Figure~\ref{fig:node_degree_distribution} features the distribution of the node degrees in the graphs constructed with each of the thresholds. As can be seen, a large portion of the nodes have degrees in the range (170,200). This indicates that the problem instances are highly similar to each other, and further explains the large cluster obtained with the clustering, and the large connected component obtained in the graphs.

Next, we compare the algorithms using all benchmark suites selected for each combination of a graph heuristic selection (DS or MIS algorithm) and each similarity thresholds 0.90, 0.95, and 0.97. Table~\ref{tab:ds_mis} presents the results obtained for comparing the algorithms on benchmark suites selected by the DS and MIS algorithms.

\begin{table}[t]
\caption{Results of the Friedman test and the Nemenyi post-hoc test for the statistical comparison of the three algorithms using the benchmark suites selected by the MIS and DS graph algorithms, for different cosine similarity measures. The numbers indicate the number of times in which no statistical significance was identified between the performance of a pair of algorithms, out of 30 independent executions of the statistical analysis, on 30 different subsets of instances produced by 30 runs of the algorithms.}
\label{tab:ds_mis}
\centering
\begin{tabular}{|r|r|r|r|r|r|r|}
  \hline
  &\multicolumn{2}{c|}{DS 0.9} &\multicolumn{2}{c|}{DS 0.95} &\multicolumn{2}{c|}{DS 0.97}\\
  \hline
 & DE & RSPSO & DE & RSPSO & DE & RSPSO \\ 
  \hline
  RSPSO & 30.00 & & 30.00 & & 30.00 &  \\ 
  CMA & 27.00 & 5.00 & 26.00 & 3.00 & 22.00 & 0.00\\   \hline
  &\multicolumn{2}{c|}{MIS 0.9} &\multicolumn{2}{c|}{MIS 0.95} &\multicolumn{2}{c|}{MIS 0.97}\\
  \hline
 & DE & RSPSO & DE & RSPSO & DE & RSPSO \\ 
  \hline
  RSPSO & 30.00 & & 30.00 & & 30.00 &  \\ 
  CMA & 27.00 & 3.00 & 30.00 & 0.00 & 24.00 & 0.00\\
   \hline
\end{tabular}
\end{table}

We need to point out that in this experiment we do not report the p-values, because we would like to trace the information if the same statistical outcome is achieved if we repeat the comparison on different benchmark suites selected by the same graph heuristic, as in a bootstrapping evaluation process. 

The results in Table~\ref{tab:ds_mis} show that the statistical outcome is reproducible when the benchmark suites are selected by the graph heuristics. For example, when the comparison is performed using problem instances selected with the DS algorithm and a similarity threshold of 0.90 used for generating the graph, in 30 out of 30 independent comparisons (i.e., comparing the algorithms on 30 benchmark suites selected by the DS algorithm with different random seeds) there is no statistical significance between the performance of DE and RSPSO. The same holds for the pair (DE, CMA), since in 27 out of 30 selected benchmark suites the same statistical outcome was achieved. Finally, the statistical outcome of the comparison between CMA and RSPSO is also robust since it shows that in 30-5=25 out of 30 comparisons a statistical significance between their performance was found (resulting in 0 in our counting process, so the end results presented in the table is 5).

From Table~\ref{tab:ds_mis} we can conclude that both graph heuristics generate a selection of representative benchmark suites that provide reproducible statistical outcomes from comparison studies. We need to point out that in the proposed approach the statistical analysis should be done several times in order to guarantee the robustness and reproducibility of the statistical outcomes.

\subsection{Generalization of the proposed heuristics}
\begin{figure}
    \centering
    \includegraphics[width=\linewidth]{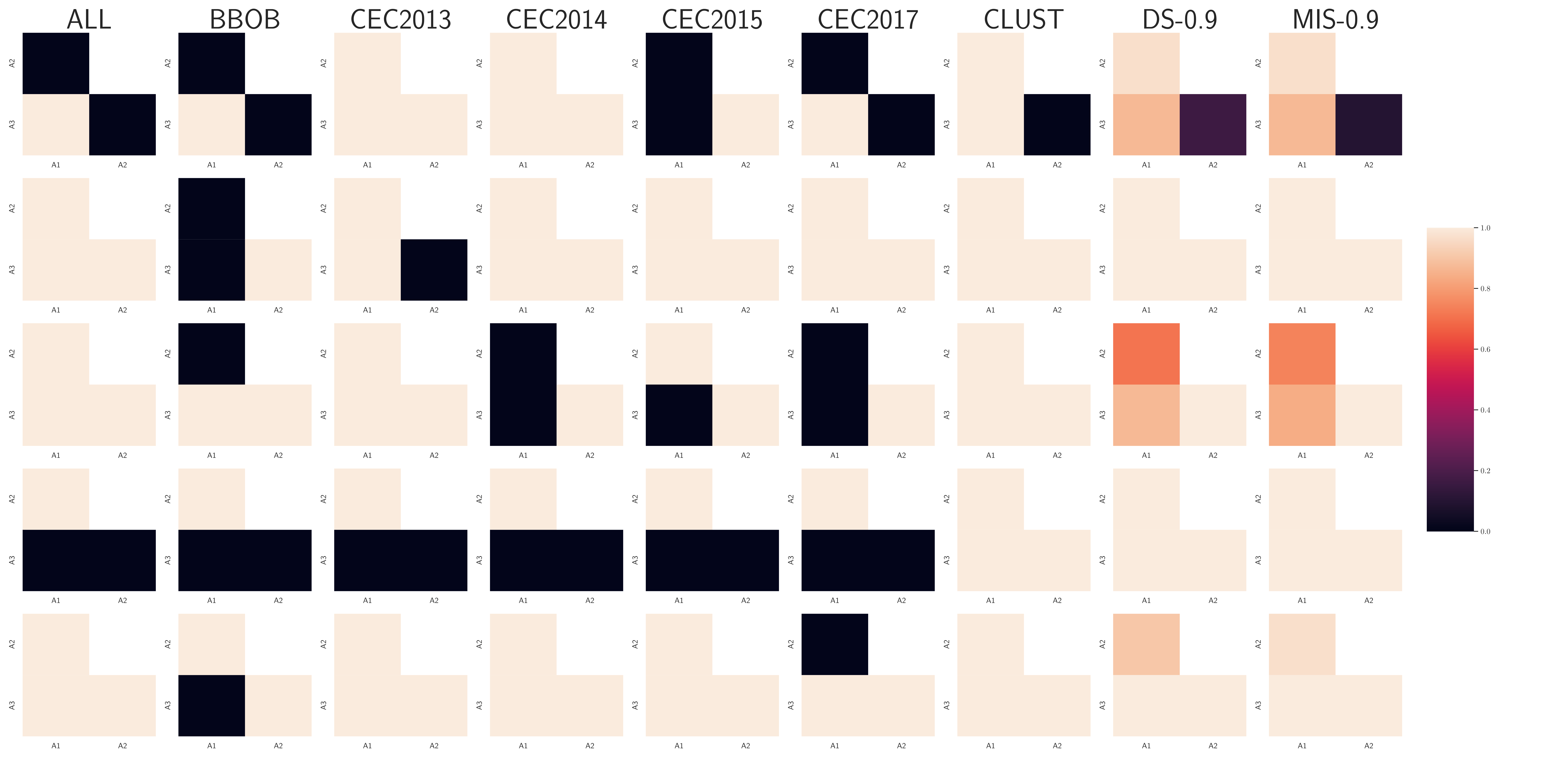}
    \caption{Visualization of the statistical outcomes of the Friedman and Nemenyi tests from the analysis on all five algorithm portfolios. Each row represents a different algorithm portfolio. The columns indicate the set of instances used for comparing the algorithms. Experiments were conducted on all benchmarks (column 1), on each benchmark separately (columns 2-6), on the instances selected by the clustering using 21 clusters (column 7) and the instances selected by the DS and MIS algorithms with a similarity threshold of 0.9 (columns 8-9). The analysis with the instances obtained with the clustering is performed 15 times, while the analysis with the instances from the DS and MIS algorithms are performed 30 times. The binary outcomes from individual runs are summed and normalized in the range [0,1] to enable a comparison with the rest of the approaches for which the analysis is executed once.
    Lighter values indicate that no statistical significance has been found between the difference in performance of the algorithms, while darker values indicate that the algorithms are statistically different.
    }
    \label{fig:portfolio_analysis}
\end{figure}

In this paper, we demonstrated the applicability of the proposed heuristics with the statistical analysis of the DE, CMA and RSPSO algorithms. To prove the generalization of the proposed heuristics, we conducted the same analysis for four additional portfolios, each containing three arbitrarily chosen optimization algorithms.
We used the same approaches for selecting the problem instances independently from the algorithm portfolio involved in the comparison. Further, we compared each algorithm portfolio and for each one we generated all tables that are presented for the portfolio presented in the paper. 
Figure~\ref{fig:portfolio_analysis} summarizes the results for all five algorithm portfolios by depicting the statistical outcomes of the pairwise comparisons of the algorithms. The full results can be found in our GitHub repository.
For the second portfolio, for which we do not report all of the tables in the paper, we can see that the outcomes obtained on BBOB and CEC2013 are different from the rest. For the third portfolio and fifth portfolio, the evaluations done on a single benchmark produce conflicting results.
In all portfolios, the evaluation done using the subsets of instances produced by the proposed heuristics produce consistent and reliable results.

\section{Conclusion}
\label{sec:conclusion}

In this paper, we propose three approaches for selecting benchmark subsets which uniformly cover the problem landscape and can provide a reliable and robust statistical analysis of algorithm performance. We show that when the statistical analysis is conducted on a single benchmark set, one can obtain different outcomes regarding the superiority of one algorithm over another, when a different already established benchmark set is used. 
The benchmark subsets produced by the three proposed approaches provide a consistent and reliable evaluation.


Our study uses the landscape features computed in~\cite{lang2021exploratory}. This work includes 118 miscellaneous functions, which we did not include in our study because no performance data was available for them. An extension of our work to include these functions is a straightforward next step, as is the coverage of a larger algorithm portfolio, and testing the sensitivity of the methodology when different similarity measures will be applied instead of cosine similarity and transformation of the ELA representations~\cite{eftimov2020linear}.

The existing optimization benchmarks are developed to understand algorithms' strengths and weaknesses, and not to achieve results that show better transferability to other problems using machine learning (ML) approaches~\cite{lacroix2019,skvorc2022transfer}. The proposed approaches can also be used to select problem instances for training ML models with better transferability to other problems in automated performance prediction, selection, or configuration. 

\begin{acks}

\noindent  This work is based on the research supported by Slovenian Research Agency: research  core  fundings  No.  P2-0098 and project  No.  N2-0239 to TE, and scholarship by the Ad Futura grant for postgraduate study to GC.

\noindent This work is also based on the research supported in part by the National Research Foundation (NRF) of South Africa.
The findings in this work is that of the authors alone, and the NRF accepts no liability whatsoever in this regard. 

\noindent Carola Doerr acknowledges financial support from the Paris Ile-de-France Region region. 

The authors acknowledge the Centre for High Performance Computing (CHPC), South Africa, for providing computational resources to this research project.

\end{acks}

\bibliographystyle{ACM-Reference-Format}
\bibliography{bib}


\begin{thebibliography}{43}


\ifx \showCODEN    \undefined \def \showCODEN     #1{\unskip}     \fi
\ifx \showDOI      \undefined \def \showDOI       #1{#1}\fi
\ifx \showISBNx    \undefined \def \showISBNx     #1{\unskip}     \fi
\ifx \showISBNxiii \undefined \def \showISBNxiii  #1{\unskip}     \fi
\ifx \showISSN     \undefined \def \showISSN      #1{\unskip}     \fi
\ifx \showLCCN     \undefined \def \showLCCN      #1{\unskip}     \fi
\ifx \shownote     \undefined \def \shownote      #1{#1}          \fi
\ifx \showarticletitle \undefined \def \showarticletitle #1{#1}   \fi
\ifx \showURL      \undefined \def \showURL       {\relax}        \fi
\providecommand\bibfield[2]{#2}
\providecommand\bibinfo[2]{#2}
\providecommand\natexlab[1]{#1}
\providecommand\showeprint[2][]{arXiv:#2}

\bibitem[\protect\citeauthoryear{Allan and Laskar}{Allan and Laskar}{1978}]%
        {definitions_dom}
\bibfield{author}{\bibinfo{person}{Robert~B. Allan} {and} \bibinfo{person}{Renu
  Laskar}.} \bibinfo{year}{1978}\natexlab{}.
\newblock \showarticletitle{On domination and independent domination numbers of
  a graph}.
\newblock \bibinfo{journal}{\emph{Discrete Mathematics}} \bibinfo{volume}{23},
  \bibinfo{number}{2} (\bibinfo{year}{1978}), \bibinfo{pages}{73--76}.
\newblock
\showISSN{0012-365X}
\urldef\tempurl%
\url{https://doi.org/10.1016/0012-365X(78)90105-X}
\showDOI{\tempurl}


\bibitem[\protect\citeauthoryear{Beachkofski and Grandhi}{Beachkofski and
  Grandhi}{2002}]%
        {iLHS}
\bibfield{author}{\bibinfo{person}{B. Beachkofski} {and} \bibinfo{person}{R.
  Grandhi}.} \bibinfo{year}{2002}\natexlab{}.
\newblock \showarticletitle{Improved {Distributed} {Hypercube} {Sampling}}.
\newblock In \bibinfo{booktitle}{\emph{43rd {AIAA}/{ASME}/{ASCE}/{AHS}/{ASC}
  {Structures}, {Structural} {Dynamics}, and {Materials} {Conference}}}.
  \bibinfo{publisher}{American Institute of Aeronautics and Astronautics}.
\newblock
\urldef\tempurl%
\url{https://doi.org/10.2514/6.2002-1274}
\showDOI{\tempurl}


\bibitem[\protect\citeauthoryear{Calvo and Santafe}{Calvo and Santafe}{2015}]%
        {scmamp}
\bibfield{author}{\bibinfo{person}{Borja Calvo} {and} \bibinfo{person}{Guzman
  Santafe}.} \bibinfo{year}{2015}\natexlab{}.
\newblock \showarticletitle{scmamp: Statistical Comparison of Multiple
  Algorithms in Multiple Problems}.
\newblock \bibinfo{journal}{\emph{The R Journal}}  \bibinfo{volume}{Accepted
  for publication} (\bibinfo{year}{2015}).
\newblock


\bibitem[\protect\citeauthoryear{Cawley and Talbot}{Cawley and Talbot}{2010}]%
        {overfitting_model_selection_bias}
\bibfield{author}{\bibinfo{person}{Gavin~C. Cawley} {and}
  \bibinfo{person}{Nicola L.~C. Talbot}.} \bibinfo{year}{2010}\natexlab{}.
\newblock \showarticletitle{On Over-fitting in Model Selection and Subsequent
  Selection Bias in Performance Evaluation}.
\newblock \bibinfo{journal}{\emph{Journal of Machine Learning Research}}
  \bibinfo{volume}{11}, \bibinfo{number}{70} (\bibinfo{year}{2010}),
  \bibinfo{pages}{2079--2107}.
\newblock
\urldef\tempurl%
\url{http://jmlr.org/papers/v11/cawley10a.html}
\showURL{%
\tempurl}


\bibitem[\protect\citeauthoryear{Christie, Brownlee, and Woodward}{Christie
  et~al\mbox{.}}{2018}]%
        {christie2018investigating}
\bibfield{author}{\bibinfo{person}{Lee~A Christie},
  \bibinfo{person}{Alexander~EI Brownlee}, {and} \bibinfo{person}{John~R
  Woodward}.} \bibinfo{year}{2018}\natexlab{}.
\newblock \showarticletitle{Investigating benchmark correlations when comparing
  algorithms with parameter tuning}. In \bibinfo{booktitle}{\emph{Proc. of the
  Genetic and Evolutionary Computation Conference Companion}}.
  \bibinfo{pages}{209--210}.
\newblock


\bibitem[\protect\citeauthoryear{Dem\v{s}ar}{Dem\v{s}ar}{2006}]%
        {demsar_statistics}
\bibfield{author}{\bibinfo{person}{Janez Dem\v{s}ar}.}
  \bibinfo{year}{2006}\natexlab{}.
\newblock \showarticletitle{Statistical Comparisons of Classifiers over
  Multiple Data Sets}.
\newblock   \bibinfo{volume}{7} (\bibinfo{date}{dec} \bibinfo{year}{2006}),
  \bibinfo{pages}{1–30}.
\newblock
\showISSN{1532-4435}


\bibitem[\protect\citeauthoryear{Eftimov, Koro{\v{s}}ec, and Seljak}{Eftimov
  et~al\mbox{.}}{2017}]%
        {eftimov2017novel}
\bibfield{author}{\bibinfo{person}{Tome Eftimov}, \bibinfo{person}{Peter
  Koro{\v{s}}ec}, {and} \bibinfo{person}{Barbara~Korou{\v{s}}i{\'c} Seljak}.}
  \bibinfo{year}{2017}\natexlab{}.
\newblock \showarticletitle{A novel approach to statistical comparison of
  meta-heuristic stochastic optimization algorithms using deep statistics}.
\newblock \bibinfo{journal}{\emph{Information Sciences}}  \bibinfo{volume}{417}
  (\bibinfo{year}{2017}), \bibinfo{pages}{186--215}.
\newblock


\bibitem[\protect\citeauthoryear{Eftimov, Popovski, Renau, Koro{\v{s}}ec, and
  Doerr}{Eftimov et~al\mbox{.}}{2020}]%
        {eftimov2020linear}
\bibfield{author}{\bibinfo{person}{Tome Eftimov}, \bibinfo{person}{Gorjan
  Popovski}, \bibinfo{person}{Quentin Renau}, \bibinfo{person}{Peter
  Koro{\v{s}}ec}, {and} \bibinfo{person}{Carola Doerr}.}
  \bibinfo{year}{2020}\natexlab{}.
\newblock \showarticletitle{Linear matrix factorization embeddings for
  single-objective optimization landscapes}. In \bibinfo{booktitle}{\emph{2020
  IEEE Symposium Series on Computational Intelligence (SSCI)}}. IEEE,
  \bibinfo{pages}{775--782}.
\newblock


\bibitem[\protect\citeauthoryear{Garc{\'\i}a, Molina, Lozano, and
  Herrera}{Garc{\'\i}a et~al\mbox{.}}{2009}]%
        {garcia2009study}
\bibfield{author}{\bibinfo{person}{Salvador Garc{\'\i}a},
  \bibinfo{person}{Daniel Molina}, \bibinfo{person}{Manuel Lozano}, {and}
  \bibinfo{person}{Francisco Herrera}.} \bibinfo{year}{2009}\natexlab{}.
\newblock \showarticletitle{A study on the use of non-parametric tests for
  analyzing the evolutionary algorithms’ behaviour: a case study on the
  CEC’2005 special session on real parameter optimization}.
\newblock \bibinfo{journal}{\emph{Journal of Heuristics}} \bibinfo{volume}{15},
  \bibinfo{number}{6} (\bibinfo{year}{2009}), \bibinfo{pages}{617--644}.
\newblock


\bibitem[\protect\citeauthoryear{Hagberg, Schult, and Swart}{Hagberg
  et~al\mbox{.}}{2008}]%
        {networkx}
\bibfield{author}{\bibinfo{person}{Aric~A. Hagberg}, \bibinfo{person}{Daniel~A.
  Schult}, {and} \bibinfo{person}{Pieter~J. Swart}.}
  \bibinfo{year}{2008}\natexlab{}.
\newblock \showarticletitle{Exploring Network Structure, Dynamics, and Function
  using NetworkX}.
\newblock  (\bibinfo{year}{2008}), \bibinfo{pages}{11 -- 15}.
\newblock


\bibitem[\protect\citeauthoryear{Hansen, Auger, Ros, Mersmann, Tu{\v{s}}ar, and
  Brockhoff}{Hansen et~al\mbox{.}}{2020}]%
        {hansen2020coco}
\bibfield{author}{\bibinfo{person}{Nikolaus Hansen}, \bibinfo{person}{Anne
  Auger}, \bibinfo{person}{Raymond Ros}, \bibinfo{person}{Olaf Mersmann},
  \bibinfo{person}{Tea Tu{\v{s}}ar}, {and} \bibinfo{person}{Dimo Brockhoff}.}
  \bibinfo{year}{2020}\natexlab{}.
\newblock \showarticletitle{COCO: A platform for comparing continuous
  optimizers in a black-box setting}.
\newblock \bibinfo{journal}{\emph{Optimization Methods and Software}}
  (\bibinfo{year}{2020}), \bibinfo{pages}{1--31}.
\newblock


\bibitem[\protect\citeauthoryear{Hansen, Finck, Ros, and Auger}{Hansen
  et~al\mbox{.}}{2009}]%
        {bbob}
\bibfield{author}{\bibinfo{person}{Nikolaus Hansen}, \bibinfo{person}{Steffen
  Finck}, \bibinfo{person}{Raymond Ros}, {and} \bibinfo{person}{Anne Auger}.}
  \bibinfo{year}{2009}\natexlab{}.
\newblock \bibinfo{booktitle}{\emph{{Real-Parameter Black-Box Optimization
  Benchmarking 2009: Noiseless Functions Definitions}}}.
\newblock \bibinfo{type}{Research Report} RR-6829.
  \bibinfo{institution}{{INRIA}}.
\newblock
\urldef\tempurl%
\url{https://hal.inria.fr/inria-00362633}
\showURL{%
\tempurl}


\bibitem[\protect\citeauthoryear{Hansen and Ostermeier}{Hansen and
  Ostermeier}{2001}]%
        {hansen2001self_adaptation_es}
\bibfield{author}{\bibinfo{person}{Nikolaus Hansen} {and}
  \bibinfo{person}{Andreas Ostermeier}.} \bibinfo{year}{2001}\natexlab{}.
\newblock \showarticletitle{Completely Derandomized Self-Adaptation in
  Evolution Strategies}.
\newblock \bibinfo{journal}{\emph{Evolutionary Computation}}
  \bibinfo{volume}{9}, \bibinfo{number}{2} (\bibinfo{year}{2001}),
  \bibinfo{pages}{159--195}.
\newblock
\urldef\tempurl%
\url{https://doi.org/10.1162/106365601750190398}
\showDOI{\tempurl}


\bibitem[\protect\citeauthoryear{Jeavons, Scott, and Xu}{Jeavons
  et~al\mbox{.}}{2016}]%
        {definitions_mis}
\bibfield{author}{\bibinfo{person}{Peter Jeavons}, \bibinfo{person}{Alex~D.
  Scott}, {and} \bibinfo{person}{Lei Xu}.} \bibinfo{year}{2016}\natexlab{}.
\newblock \showarticletitle{Feedback from nature: simple randomised distributed
  algorithms for maximal independent set selection and greedy colouring}.
\newblock \bibinfo{journal}{\emph{Distributed Computing}}  \bibinfo{volume}{29}
  (\bibinfo{year}{2016}), \bibinfo{pages}{377--393}.
\newblock


\bibitem[\protect\citeauthoryear{Kennedy and Eberhart}{Kennedy and
  Eberhart}{1995}]%
        {kennedy1995pso}
\bibfield{author}{\bibinfo{person}{J. Kennedy} {and} \bibinfo{person}{R.
  Eberhart}.} \bibinfo{year}{1995}\natexlab{}.
\newblock \showarticletitle{Particle swarm optimization}. In
  \bibinfo{booktitle}{\emph{Proc. of ICNN'95 - International Conference on
  Neural Networks}}, Vol.~\bibinfo{volume}{4}. \bibinfo{pages}{1942--1948
  vol.4}.
\newblock
\urldef\tempurl%
\url{https://doi.org/10.1109/ICNN.1995.488968}
\showDOI{\tempurl}


\bibitem[\protect\citeauthoryear{Kerschke, Preuss, Wessing, and
  Trautmann}{Kerschke et~al\mbox{.}}{2015}]%
        {kerschke_detecting_2015}
\bibfield{author}{\bibinfo{person}{P. Kerschke}, \bibinfo{person}{M. Preuss},
  \bibinfo{person}{S. Wessing}, {and} \bibinfo{person}{H. Trautmann}.}
  \bibinfo{year}{2015}\natexlab{}.
\newblock \showarticletitle{Detecting {Funnel} {Structures} by {Means} of
  {Exploratory} {Landscape} {Analysis}}. In \bibinfo{booktitle}{\emph{Proc. of
  Genetic and Evolutionary Computation Conference (GECCO'15)}}.
  \bibinfo{publisher}{ACM}, \bibinfo{pages}{265--272}.
\newblock
\showISBNx{978-1-4503-3472-3}
\urldef\tempurl%
\url{https://doi.org/10.1145/2739480.2754642}
\showDOI{\tempurl}


\bibitem[\protect\citeauthoryear{Kerschke and Trautmann}{Kerschke and
  Trautmann}{2019}]%
        {KerschkeT2019flacco}
\bibfield{author}{\bibinfo{person}{Pascal Kerschke} {and}
  \bibinfo{person}{Heike Trautmann}.} \bibinfo{year}{2019}\natexlab{}.
\newblock \showarticletitle{Comprehensive Feature-Based Landscape Analysis of
  Continuous and Constrained Optimization Problems Using the R-package flacco}.
\newblock In \bibinfo{booktitle}{\emph{Applications in Statistical Computing --
  From Music Data Analysis to Industrial Quality Improvement}},
  \bibfield{editor}{\bibinfo{person}{Nadja Bauer}, \bibinfo{person}{Katja
  Ickstadt}, \bibinfo{person}{Karsten L{\"u}bke}, \bibinfo{person}{Gero
  Szepannek}, \bibinfo{person}{Heike Trautmann}, {and}
  \bibinfo{person}{Maurizio Vichi}} (Eds.). \bibinfo{publisher}{Springer},
  \bibinfo{pages}{93~--~123}.
\newblock
\urldef\tempurl%
\url{https://doi.org/10.1007/978-3-030-25147-5_7}
\showDOI{\tempurl}


\bibitem[\protect\citeauthoryear{Lacroix and McCall}{Lacroix and
  McCall}{2019}]%
        {lacroix2019}
\bibfield{author}{\bibinfo{person}{Benjamin Lacroix} {and}
  \bibinfo{person}{John McCall}.} \bibinfo{year}{2019}\natexlab{}.
\newblock \showarticletitle{Limitations of Benchmark Sets and Landscape
  Features for Algorithm Selection and Performance Prediction}. In
  \bibinfo{booktitle}{\emph{Proc. of Genetic and Evolutionary Computation
  (GECCO'19, Companion)}}. \bibinfo{publisher}{ACM},
  \bibinfo{pages}{261–262}.
\newblock
\urldef\tempurl%
\url{https://doi.org/10.1145/3319619.3322051}
\showDOI{\tempurl}


\bibitem[\protect\citeauthoryear{Lang and Engelbrecht}{Lang and
  Engelbrecht}{2021}]%
        {lang2021exploratory}
\bibfield{author}{\bibinfo{person}{Ryan~Dieter Lang} {and}
  \bibinfo{person}{Andries~Petrus Engelbrecht}.}
  \bibinfo{year}{2021}\natexlab{}.
\newblock \showarticletitle{An Exploratory Landscape Analysis-Based Benchmark
  Suite}.
\newblock \bibinfo{journal}{\emph{Algorithms}} \bibinfo{volume}{14},
  \bibinfo{number}{3} (\bibinfo{year}{2021}).
\newblock
\showISSN{1999-4893}
\urldef\tempurl%
\url{https://doi.org/10.3390/a14030078}
\showDOI{\tempurl}


\bibitem[\protect\citeauthoryear{Liang, Qu, and Suganthan}{Liang
  et~al\mbox{.}}{2013a}]%
        {cec2014}
\bibfield{author}{\bibinfo{person}{J.J. Liang}, \bibinfo{person}{B. Qu}, {and}
  \bibinfo{person}{Ponnuthurai Suganthan}.} \bibinfo{year}{2013}\natexlab{a}.
\newblock \showarticletitle{Problem definitions and evaluation criteria for the
  CEC 2014 special session and competition on single objective real-parameter
  numerical optimization}.
\newblock \bibinfo{journal}{\emph{Computational Intelligence Laboratory,
  Zhengzhou University, Zhengzhou China and Technical Report, Nanyang
  Technological University, Singapore.}} (\bibinfo{date}{12}
  \bibinfo{year}{2013}).
\newblock


\bibitem[\protect\citeauthoryear{Liang, Qu, Suganthan, and Chen}{Liang
  et~al\mbox{.}}{2014}]%
        {cec2015}
\bibfield{author}{\bibinfo{person}{J.J. Liang}, \bibinfo{person}{B.Y. Qu},
  \bibinfo{person}{P.N. Suganthan}, {and} \bibinfo{person}{Q. Chen}.}
  \bibinfo{year}{2014}\natexlab{}.
\newblock \showarticletitle{Problem definitions and evaluation criteria for the
  CEC 2015 competition on learning-based real-parameter single objective
  optimization}.
\newblock \bibinfo{journal}{\emph{Computational Intelligence Laboratory,
  Zhengzhou University, Zhengzhou China and Technical Report, Nanyang
  Technological University, Singapore.}} (\bibinfo{year}{2014}).
\newblock


\bibitem[\protect\citeauthoryear{Liang, Qu, Suganthan, and
  Hernández-Díaz}{Liang et~al\mbox{.}}{2013b}]%
        {cec2013}
\bibfield{author}{\bibinfo{person}{J.J. Liang}, \bibinfo{person}{B. Qu},
  \bibinfo{person}{Ponnuthurai Suganthan}, {and} \bibinfo{person}{Alfredo
  Hernández-Díaz}.} \bibinfo{year}{2013}\natexlab{b}.
\newblock \showarticletitle{Problem Definitions and Evaluation Criteria for the
  CEC 2013 Special Session on Real-Parameter Optimization}.
\newblock \bibinfo{journal}{\emph{Computational Intelligence Laboratory,
  Zhengzhou University, Zhengzhou China and Technical Report, Nanyang
  Technological University, Singapore.}} (\bibinfo{date}{01}
  \bibinfo{year}{2013}).
\newblock


\bibitem[\protect\citeauthoryear{Lunacek and Whitley}{Lunacek and
  Whitley}{2006}]%
        {lunacek_dispersion_2006}
\bibfield{author}{\bibinfo{person}{M. Lunacek} {and} \bibinfo{person}{D.
  Whitley}.} \bibinfo{year}{2006}\natexlab{}.
\newblock \showarticletitle{The dispersion metric and the {CMA} evolution
  strategy}. In \bibinfo{booktitle}{\emph{Proc. of Genetic and Evolutionary
  Computation Conference (GECCO'06)}}. \bibinfo{publisher}{ACM},
  \bibinfo{pages}{477}.
\newblock
\showISBNx{978-1-59593-186-3}
\urldef\tempurl%
\url{https://doi.org/10.1145/1143997.1144085}
\showDOI{\tempurl}


\bibitem[\protect\citeauthoryear{Mersmann, Bischl, Trautmann, Preuss, Weihs,
  and Rudolph}{Mersmann et~al\mbox{.}}{2011}]%
        {mersmann2011exploratory}
\bibfield{author}{\bibinfo{person}{Olaf Mersmann}, \bibinfo{person}{Bernd
  Bischl}, \bibinfo{person}{Heike Trautmann}, \bibinfo{person}{Mike Preuss},
  \bibinfo{person}{Claus Weihs}, {and} \bibinfo{person}{G{\"u}nter Rudolph}.}
  \bibinfo{year}{2011}\natexlab{}.
\newblock \showarticletitle{Exploratory landscape analysis}. In
  \bibinfo{booktitle}{\emph{Proc. of Genetic and Evolutionary Computation
  Conference (GECCO'11)}}. \bibinfo{publisher}{ACM}, \bibinfo{pages}{829--836}.
\newblock
\urldef\tempurl%
\url{https://doi.org/10.1145/2001576.2001690}
\showDOI{\tempurl}


\bibitem[\protect\citeauthoryear{Meunier, Rakotoarison, Wong, Roziere, Rapin,
  Teytaud, Moreau, and Doerr}{Meunier et~al\mbox{.}}{2021}]%
        {meunier2021black}
\bibfield{author}{\bibinfo{person}{Laurent Meunier},
  \bibinfo{person}{Herilalaina Rakotoarison}, \bibinfo{person}{Pak~Kan Wong},
  \bibinfo{person}{Baptiste Roziere}, \bibinfo{person}{Jeremy Rapin},
  \bibinfo{person}{Olivier Teytaud}, \bibinfo{person}{Antoine Moreau}, {and}
  \bibinfo{person}{Carola Doerr}.} \bibinfo{year}{2021}\natexlab{}.
\newblock \showarticletitle{Black-box optimization revisited: Improving
  algorithm selection wizards through massive benchmarking}.
\newblock \bibinfo{journal}{\emph{IEEE Transactions on Evolutionary
  Computation}} (\bibinfo{year}{2021}).
\newblock


\bibitem[\protect\citeauthoryear{Muñoz, Kirley, and Halgamuge}{Muñoz
  et~al\mbox{.}}{2015}]%
        {munoz_exploratory_2015}
\bibfield{author}{\bibinfo{person}{M.A. Muñoz}, \bibinfo{person}{M. Kirley},
  {and} \bibinfo{person}{S.K. Halgamuge}.} \bibinfo{year}{2015}\natexlab{}.
\newblock \showarticletitle{Exploratory {Landscape} {Analysis} of {Continuous}
  {Space} {Optimization} {Problems} {Using} {Information} {Content}}.
\newblock \bibinfo{journal}{\emph{IEEE Transactions on Evolutionary
  Computation}} \bibinfo{volume}{19}, \bibinfo{number}{1}
  (\bibinfo{year}{2015}), \bibinfo{pages}{74--87}.
\newblock
\showISSN{1089-778X}
\urldef\tempurl%
\url{https://doi.org/10.1109/TEVC.2014.2302006}
\showDOI{\tempurl}


\bibitem[\protect\citeauthoryear{Muñoz and Smith-Miles}{Muñoz and
  Smith-Miles}{2020}]%
        {munoz_generating_new_instances}
\bibfield{author}{\bibinfo{person}{Mario~A. Muñoz} {and} \bibinfo{person}{Kate
  Smith-Miles}.} \bibinfo{year}{2020}\natexlab{}.
\newblock \showarticletitle{{Generating New Space-Filling Test Instances for
  Continuous Black-Box Optimization}}.
\newblock \bibinfo{journal}{\emph{Evolutionary Computation}}
  \bibinfo{volume}{28}, \bibinfo{number}{3} (\bibinfo{date}{09}
  \bibinfo{year}{2020}), \bibinfo{pages}{379--404}.
\newblock
\showISSN{1063-6560}
\urldef\tempurl%
\url{https://doi.org/10.1162/evco_a_00262}
\showDOI{\tempurl}
\showeprint{https://direct.mit.edu/evco/article-pdf/28/3/379/1858988/evco\_a\_00262.pdf}


\bibitem[\protect\citeauthoryear{Nemenyi}{Nemenyi}{1963}]%
        {nemenyi}
\bibfield{author}{\bibinfo{person}{P. Nemenyi}.}
  \bibinfo{year}{1963}\natexlab{}.
\newblock \bibinfo{booktitle}{\emph{Distribution-free Multiple Comparisons}}.
\newblock \bibinfo{publisher}{Princeton University}.
\newblock
\urldef\tempurl%
\url{https://books.google.si/books?id=nhDMtgAACAAJ}
\showURL{%
\tempurl}


\bibitem[\protect\citeauthoryear{Pedregosa, Varoquaux, Gramfort, Michel,
  Thirion, Grisel, Blondel, Prettenhofer, Weiss, Dubourg, Vanderplas, Passos,
  Cournapeau, Brucher, Perrot, and Duchesnay}{Pedregosa et~al\mbox{.}}{2011}]%
        {scikit-learn}
\bibfield{author}{\bibinfo{person}{F. Pedregosa}, \bibinfo{person}{G.
  Varoquaux}, \bibinfo{person}{A. Gramfort}, \bibinfo{person}{V. Michel},
  \bibinfo{person}{B. Thirion}, \bibinfo{person}{O. Grisel},
  \bibinfo{person}{M. Blondel}, \bibinfo{person}{P. Prettenhofer},
  \bibinfo{person}{R. Weiss}, \bibinfo{person}{V. Dubourg}, \bibinfo{person}{J.
  Vanderplas}, \bibinfo{person}{A. Passos}, \bibinfo{person}{D. Cournapeau},
  \bibinfo{person}{M. Brucher}, \bibinfo{person}{M. Perrot}, {and}
  \bibinfo{person}{E. Duchesnay}.} \bibinfo{year}{2011}\natexlab{}.
\newblock \showarticletitle{Scikit-learn: Machine Learning in {P}ython}.
\newblock \bibinfo{journal}{\emph{Journal of Machine Learning Research}}
  \bibinfo{volume}{12} (\bibinfo{year}{2011}), \bibinfo{pages}{2825--2830}.
\newblock


\bibitem[\protect\citeauthoryear{Pohlert}{Pohlert}{2014}]%
        {pmcmr}
\bibfield{author}{\bibinfo{person}{Thorsten Pohlert}.}
  \bibinfo{year}{2014}\natexlab{}.
\newblock \bibinfo{booktitle}{\emph{The Pairwise Multiple Comparison of Mean
  Ranks Package (PMCMR)}}.
\newblock
\urldef\tempurl%
\url{https://CRAN.R-project.org/package=PMCMR}
\showURL{%
\tempurl}
\newblock
\shownote{R package}.


\bibitem[\protect\citeauthoryear{Rapin and Teytaud}{Rapin and Teytaud}{2018}]%
        {nevergrad_github}
\bibfield{author}{\bibinfo{person}{J. Rapin} {and} \bibinfo{person}{O.
  Teytaud}.} \bibinfo{year}{2018}\natexlab{}.
\newblock \bibinfo{title}{{Nevergrad - A gradient-free optimization platform}}.
\newblock
  \bibinfo{howpublished}{\url{https://GitHub.com/FacebookResearch/Nevergrad}}.
\newblock


\bibitem[\protect\citeauthoryear{Shirkhorshidi, Aghabozorgi, and
  Wah}{Shirkhorshidi et~al\mbox{.}}{2015}]%
        {shirkhorshidi2015comparison}
\bibfield{author}{\bibinfo{person}{Ali~Seyed Shirkhorshidi},
  \bibinfo{person}{Saeed Aghabozorgi}, {and} \bibinfo{person}{Teh~Ying Wah}.}
  \bibinfo{year}{2015}\natexlab{}.
\newblock \showarticletitle{A comparison study on similarity and dissimilarity
  measures in clustering continuous data}.
\newblock \bibinfo{journal}{\emph{PloS one}} \bibinfo{volume}{10},
  \bibinfo{number}{12} (\bibinfo{year}{2015}), \bibinfo{pages}{e0144059}.
\newblock


\bibitem[\protect\citeauthoryear{{\v{S}}kvorc, Eftimov, and
  Koro{\v{s}}ec}{{\v{S}}kvorc et~al\mbox{.}}{2020a}]%
        {isa_optimization}
\bibfield{author}{\bibinfo{person}{Urban {\v{S}}kvorc}, \bibinfo{person}{Tome
  Eftimov}, {and} \bibinfo{person}{Peter Koro{\v{s}}ec}.}
  \bibinfo{year}{2020}\natexlab{a}.
\newblock \showarticletitle{Understanding the problem space in single-objective
  numerical optimization using exploratory landscape analysis}.
\newblock \bibinfo{journal}{\emph{Applied Soft Computing}}
  \bibinfo{volume}{90} (\bibinfo{date}{May} \bibinfo{year}{2020}),
  \bibinfo{pages}{106138}.
\newblock
\urldef\tempurl%
\url{https://doi.org/10.1016/j.asoc.2020.106138}
\showDOI{\tempurl}


\bibitem[\protect\citeauthoryear{{\v{S}}kvorc, Eftimov, and
  Koro{\v{s}}ec}{{\v{S}}kvorc et~al\mbox{.}}{2020b}]%
        {vskvorc2020understanding}
\bibfield{author}{\bibinfo{person}{Urban {\v{S}}kvorc}, \bibinfo{person}{Tome
  Eftimov}, {and} \bibinfo{person}{Peter Koro{\v{s}}ec}.}
  \bibinfo{year}{2020}\natexlab{b}.
\newblock \showarticletitle{Understanding the problem space in single-objective
  numerical optimization using exploratory landscape analysis}.
\newblock \bibinfo{journal}{\emph{Applied Soft Computing}}
  \bibinfo{volume}{90} (\bibinfo{year}{2020}), \bibinfo{pages}{106138}.
\newblock


\bibitem[\protect\citeauthoryear{Skvorc, Eftimov, and Korosec}{Skvorc
  et~al\mbox{.}}{2021}]%
        {urban_complementary_analysis}
\bibfield{author}{\bibinfo{person}{Urban Skvorc}, \bibinfo{person}{Tome
  Eftimov}, {and} \bibinfo{person}{Peter Korosec}.}
  \bibinfo{year}{2021}\natexlab{}.
\newblock \showarticletitle{A Complementarity Analysis of the {COCO} Benchmark
  Problems and Artificially Generated Problems}.
\newblock \bibinfo{journal}{\emph{CoRR}}  \bibinfo{volume}{abs/2104.13060}
  (\bibinfo{year}{2021}).
\newblock
\showeprint[arXiv]{2104.13060}
\urldef\tempurl%
\url{https://arxiv.org/abs/2104.13060}
\showURL{%
\tempurl}


\bibitem[\protect\citeauthoryear{Smith-Miles, Baatar, Wreford, and
  Lewis}{Smith-Miles et~al\mbox{.}}{2014}]%
        {smith2014towards}
\bibfield{author}{\bibinfo{person}{Kate Smith-Miles},
  \bibinfo{person}{Davaatseren Baatar}, \bibinfo{person}{Brendan Wreford},
  {and} \bibinfo{person}{Rhyd Lewis}.} \bibinfo{year}{2014}\natexlab{}.
\newblock \showarticletitle{Towards objective measures of algorithm performance
  across instance space}.
\newblock \bibinfo{journal}{\emph{Computers \& Operations Research}}
  \bibinfo{volume}{45} (\bibinfo{year}{2014}), \bibinfo{pages}{12--24}.
\newblock


\bibitem[\protect\citeauthoryear{Storn and Price}{Storn and Price}{1997}]%
        {storn1997differential}
\bibfield{author}{\bibinfo{person}{Rainer Storn} {and} \bibinfo{person}{Kenneth
  Price}.} \bibinfo{year}{1997}\natexlab{}.
\newblock \showarticletitle{{Differential Evolution – A Simple and Efficient
  Heuristic for global Optimization over Continuous Spaces}}.
\newblock \bibinfo{journal}{\emph{Journal of Global Optimization}}
  \bibinfo{volume}{11}, \bibinfo{number}{4} (\bibinfo{year}{1997}),
  \bibinfo{pages}{341--359}.
\newblock
\urldef\tempurl%
\url{https://doi.org/10.1023/A:1008202821328}
\showDOI{\tempurl}


\bibitem[\protect\citeauthoryear{Weise, Chen, Li, and Wu}{Weise
  et~al\mbox{.}}{2020}]%
        {Weise20ASOCwmodel}
\bibfield{author}{\bibinfo{person}{Thomas Weise}, \bibinfo{person}{Yan Chen},
  \bibinfo{person}{Xinlu Li}, {and} \bibinfo{person}{Zhize Wu}.}
  \bibinfo{year}{2020}\natexlab{}.
\newblock \showarticletitle{Selecting a diverse set of benchmark instances from
  a tunable model problem for black-box discrete optimization algorithms}.
\newblock \bibinfo{journal}{\emph{Appl. Soft Comput.}}  \bibinfo{volume}{92}
  (\bibinfo{year}{2020}), \bibinfo{pages}{106269}.
\newblock
\urldef\tempurl%
\url{https://doi.org/10.1016/j.asoc.2020.106269}
\showDOI{\tempurl}


\bibitem[\protect\citeauthoryear{Wu, Mallipeddi, and Suganthan}{Wu
  et~al\mbox{.}}{2016}]%
        {cec2017}
\bibfield{author}{\bibinfo{person}{Guohua Wu}, \bibinfo{person}{Rammohan
  Mallipeddi}, {and} \bibinfo{person}{Ponnuthurai Suganthan}.}
  \bibinfo{year}{2016}\natexlab{}.
\newblock \showarticletitle{Problem Definitions and Evaluation Criteria for the
  CEC 2017 Competition and Special Session on Constrained Single Objective
  Real-Parameter Optimization}.
\newblock \bibinfo{journal}{\emph{Computational Intelligence Laboratory,
  Zhengzhou University, Zhengzhou China and Technical Report, Nanyang
  Technological University, Singapore.}} (\bibinfo{date}{10}
  \bibinfo{year}{2016}).
\newblock


\bibitem[\protect\citeauthoryear{Yap, Munoz, Smith-Miles, and Liefooghe}{Yap
  et~al\mbox{.}}{2020}]%
        {isa_optimization_moo}
\bibfield{author}{\bibinfo{person}{Estefania Yap}, \bibinfo{person}{Mario~A.
  Munoz}, \bibinfo{person}{Kate Smith-Miles}, {and} \bibinfo{person}{Arnaud
  Liefooghe}.} \bibinfo{year}{2020}\natexlab{}.
\newblock \showarticletitle{Instance Space Analysis of Combinatorial
  Multi-objective Optimization Problems}. In \bibinfo{booktitle}{\emph{2020
  {IEEE} Congress on Evolutionary Computation ({CEC})}}.
  \bibinfo{publisher}{{IEEE}}.
\newblock
\urldef\tempurl%
\url{https://doi.org/10.1109/cec48606.2020.9185664}
\showDOI{\tempurl}


\bibitem[\protect\citeauthoryear{Yuan and Gallagher}{Yuan and
  Gallagher}{2004}]%
        {racing_problem_selection}
\bibfield{author}{\bibinfo{person}{Bo Yuan} {and} \bibinfo{person}{Marcus
  Gallagher}.} \bibinfo{year}{2004}\natexlab{}.
\newblock \showarticletitle{Statistical Racing Techniques for Improved
  Empirical Evaluation of Evolutionary Algorithms}. In
  \bibinfo{booktitle}{\emph{Parallel Problem Solving from Nature - PPSN VIII}},
  \bibfield{editor}{\bibinfo{person}{Xin Yao}, \bibinfo{person}{Edmund~K.
  Burke}, \bibinfo{person}{Jos{\'e}~A. Lozano}, \bibinfo{person}{Jim Smith},
  \bibinfo{person}{Juan~Juli{\'a}n Merelo-Guerv{\'o}s},
  \bibinfo{person}{John~A. Bullinaria}, \bibinfo{person}{Jonathan~E. Rowe},
  \bibinfo{person}{Peter Ti{\v{n}}o}, \bibinfo{person}{Ata Kab{\'a}n}, {and}
  \bibinfo{person}{Hans-Paul Schwefel}} (Eds.). \bibinfo{publisher}{Springer
  Berlin Heidelberg}, \bibinfo{address}{Berlin, Heidelberg},
  \bibinfo{pages}{172--181}.
\newblock
\showISBNx{978-3-540-30217-9}


\bibitem[\protect\citeauthoryear{Zhang and Halgamuge}{Zhang and
  Halgamuge}{2019}]%
        {zhang2019similarity}
\bibfield{author}{\bibinfo{person}{Yong-Wei Zhang} {and}
  \bibinfo{person}{Saman~K Halgamuge}.} \bibinfo{year}{2019}\natexlab{}.
\newblock \showarticletitle{Similarity of Continuous Optimization Problems from
  the Algorithm Performance Perspective}. In \bibinfo{booktitle}{\emph{2019
  IEEE Congress on Evolutionary Computation (CEC)}}. IEEE,
  \bibinfo{pages}{2949--2957}.
\newblock


\bibitem[\protect\citeauthoryear{Škvorc, Eftimov, and Korošec}{Škvorc
  et~al\mbox{.}}{2022}]%
        {skvorc2022transfer}
\bibfield{author}{\bibinfo{person}{Urban Škvorc}, \bibinfo{person}{Tome
  Eftimov}, {and} \bibinfo{person}{Peter Korošec}.}
  \bibinfo{year}{2022}\natexlab{}.
\newblock \showarticletitle{Transfer Learning Analysis of Multi-Class
  Classification for Landscape-Aware Algorithm Selection}.
\newblock \bibinfo{journal}{\emph{Mathematics}} \bibinfo{volume}{10},
  \bibinfo{number}{3} (\bibinfo{year}{2022}).
\newblock
\showISSN{2227-7390}
\urldef\tempurl%
\url{https://doi.org/10.3390/math10030432}
\showDOI{\tempurl}


\end{thebibliography}

\end{document}